\documentclass[runningheads]{llncs}
\usepackage{graphicx}

\usepackage{tikz}
\usepackage{comment}
\usepackage{amsmath,amssymb} 
\usepackage{color}

\usepackage[accsupp]{axessibility}  

\usepackage{float}
\usepackage{multirow}
\usepackage{booktabs} 
\usepackage{pifont}
\usepackage{makecell}
\usepackage{alphalph}
\usepackage{hyperref}

\newcolumntype{I}{!{\vrule width 1.2pt}}
\newlength\savedwidth

\newcommand{\tabincell}[2]{\begin{tabular}{@{}#1@{}}#2\end{tabular}}

\newcommand{\xmark}{\ding{55}}
\newcommand{\rmark}{\ding{52}}

\usepackage[capitalize]{cleveref}
\crefname{section}{Sec.}{Secs.}
\Crefname{section}{Section}{Sections}
\Crefname{table}{Table}{Tables}
\crefname{table}{Table}{Tabs.}

\begin{document}
\pagestyle{headings}
\mainmatter
\def\ECCVSubNumber{100}  

\title{An End-to-End Transformer Model for Crowd Localization} 

\titlerunning{An End-to-End Transformer Model for Crowd Localization}
%
\author{
Dingkang Liang\inst{1} \and
Wei Xu\inst{2} \and
Xiang Bai\inst{1,\dag}}
\authorrunning{D. Liang et al.}
%
\institute{Huazhong University of Science and Technology, Wuhan {\rm 430074}, China\\
\email{\{dkliang,xbai\}@hust.edu.cn} \and
Beijing University of Posts and Telecommunications, Beijing {\rm 100876}, China\\
\email{xuwei2020@bupt.edu.cn}\\
\inst{\dag}Corresponding Author}

\maketitle
\setcounter{footnote}{0}

\begin{abstract}
Crowd localization, predicting head positions, is a more practical and high-level task than simply counting. Existing methods employ pseudo-bounding boxes or pre-designed localization maps, relying on complex post-processing to obtain the head positions. In this paper, we propose an elegant, end-to-end \textbf{C}rowd \textbf{L}ocalization \textbf{TR}ansformer named CLTR that solves the task in the regression-based paradigm. The proposed method views the crowd localization as a direct set prediction problem, taking extracted features and trainable embeddings as input of the transformer-decoder. To reduce the ambiguous points and generate more reasonable matching results, we introduce a KMO-based Hungarian matcher, which adopts the nearby context as the auxiliary matching cost. Extensive experiments conducted on five datasets in various data settings show the effectiveness of our method. In particular, the proposed method achieves the best localization performance on the NWPU-Crowd, UCF-QNRF, and ShanghaiTech Part A datasets. \let\thefootnote\relax\footnotetext{Project page at \href{https://dk-liang.github.io/CLTR/}{https://dk-liang.github.io/CLTR/}}
\keywords{Crowd localization, crowd counting, transformer}
\end{abstract}

\section{Introduction}

Crowd localization, a fundamental subtask of crowd analysis, aims to provide the location of each instance. Here, the location means the center points of heads because annotating the bounding box for each head is expensive and laborious in dense scenes. Thus, most crowd datasets only provide point-level annotations. A powerful crowd localization algorithm can give great potential for similar tasks,~\textit{e.g.,} crowd tracking~\cite{wen2021detection}, object counting~\cite{li2018csrnet,du2020visdrone}, and object localization~\cite{xu2019autoscale,chen2021cell}.


\begin{figure}[t]
	\begin{center}
		\includegraphics[width=0.96\linewidth]{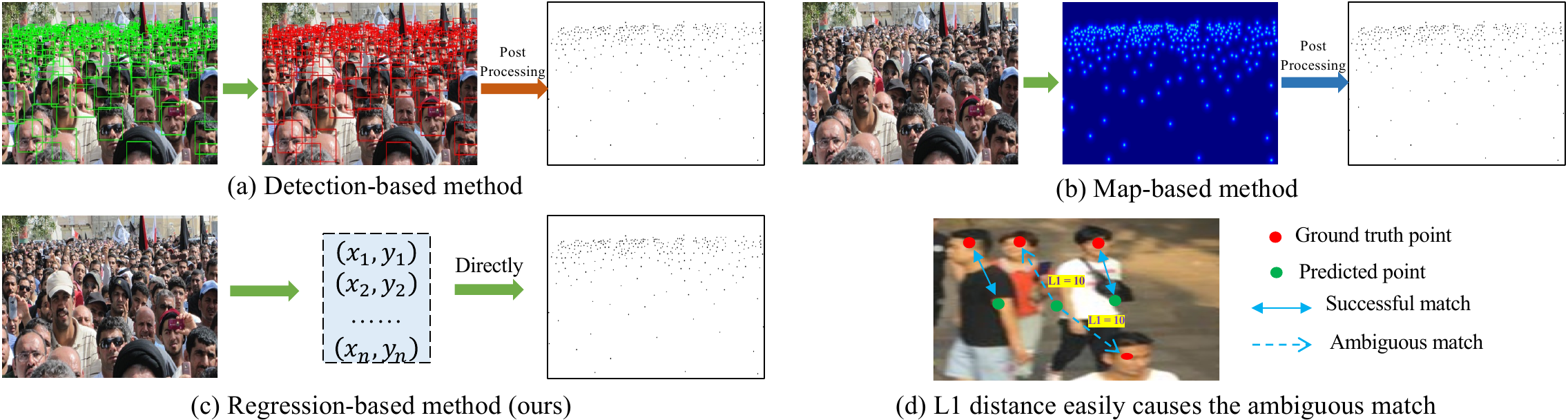}
	\end{center}
	\caption{(a) Detection-based methods, using the predefined pseudo GT bounding boxes. (b) Map-based methods, using high-resolution representation and complicated post-processing. (c) Regression-based methods, mapping the input images to the point coordinates directly. (d) Due to lack of context, the $L1$ distance easily causes the ambiguous match pair.}
	\label{fig:intro}
\end{figure}

The mainstream crowd localization methods can be generally categorized into detection-based (Fig.~\ref{fig:intro}(a)) and map-based (Fig.~\ref{fig:intro}(b)) methods. The detection-based methods~\cite{sam2020locate,liu2019point} utilize nearest-neighbor head distances to initialize the pseudo ground truth (GT) bounding boxes. However, these detection-based methods can not report satisfactory performance. Moreover, the heuristic non-maximum suppression (NMS) is used to remove the negative predictions. Most crowd localization methods~\cite{abousamra2020localization,xu2019autoscale} are map-based because it has relatively higher localization accuracy. Nevertheless, the map-based methods still suffer some inevitable problems. For instance, complex multi-scale representation is necessary to generate sharp maps. Besides, they adopt non-differentiable post-processing (\textit{e.g.,} find-maxima) to extract the location, which precludes end-to-end training.

In contrast, the regression-based methods, directly regressing the coordinates, are more straightforward than the detection-based and map-based methods, as shown in Fig.~\ref{fig:intro}(c). The benefits of regression-based can be summarized as two folds. (1) It is simple, elegant, and end-to-end trainable since it does not need pre-processing (\textit{e.g.,} pseudo GT boxes or maps generation) and post-processing (\textit{e.g.,} NMS or find-maxima). (2) It does not rely on complex multi-scale fusion mechanisms to generate high-quality feature maps.

Recently, we have witnessed the rise of Transformer~\cite{carion2020end} in computer vision. A pioneer is DETR~\cite{carion2020end}, an end-to-end trainable detector without NMS, which models the relations of the object queries and context via Transformer and achieves competitive performance only using a single-level feature map. This simple and effective detection method gives rise to a question: \textit{can crowd localization be solved with such a simple model as well?}

Our answer is: ``Yes, such a framework can be applied to crowd localization." 
Indeed, it is nothing special to directly apply the DETR-based pipeline in crowd localization. However, crowd localization is quite different from object detection. DETR shows terrible performance in the crowd localization task, attributed to the intrinsic limitation of the matcher. Specifically, the key component in DETR is the $L1$-based Hungarian matcher, which measures $L1$ distance of bounding box with class confidence to match the prediction-GT bounding box pairs, showing superior performance in object detection. However, no bounding box is given in crowd datasets, and more importantly, for crowd localization, $L1$ distance easily gives rise to ambiguous matching in the point-to-point pairs (\textit{i.e.}, a point that can belong to multiple \textit{gts} simultaneously as shown in Fig.~\ref{fig:intro}(d)). The main reasons are two-fold: (1) The $L1$-based Hungarian is a local view without context. (2) Different from the object detection, the crowd images only contain one category (heads), and the dense heads usually have similar textures, reporting close confidence, confusing the matcher. 
To this end, we introduce the $k$-nearest neighbors (KNN) matching objective named KMO as an auxiliary matching cost. The KMO-based Hungarian considers the context from nearby heads, which helps to reduce the ambiguous points and generate more reasonable matching results.

In summary, the main contributions of this paper are two-fold: 1) We propose an end-to-end \textbf{C}rowd \textbf{L}ocalization \textbf{TR}ansformer framework named CLTR, which formulates the crowd localization as a point set prediction task. CLTR significantly simplifies the crowd localization pipeline by removing pre-processing and post-processing. 2) We introduce the KMO-based Hungarian bipartite matching, which takes the context from nearby heads as an auxiliary matching cost. As a result, the matcher can effectively reduce the ambiguous points and generate more reasonable matching results. 

Extensive experiments are carried out on five challenge datasets, and significant improvements from KMO-based Hungarian matcher indicate its effectiveness. In particular, just with a single-scale and low-resolution ($\frac{1}{32}$ of input images) feature map, CLTR can achieve state-of-the-art or highly competitive localization performance.

\section{Related Works}

\subsection{Detection-based methods}
The detection-based methods~\cite{sam2020locate,liu2019point,wang2021self} mainly follow the pipeline of Faster RCNN~\cite{ren2015faster}. Specifically, PSDDN~\cite{liu2019point} utilizes the nearest neighbor distance to initialize the pseudo bounding boxes and update the pseudo boxes by choosing smaller predicted boxes in the training phase. LSC-CNN~\cite{sam2020locate} also uses a similar mechanism to generate the pseudo bounding boxes and propose a new winner-take-all loss for better training at higher resolutions. These methods~\cite{sam2020locate,liu2019point,wang2021self} usually use NMS to filter the predicted boxes, which is not end-to-end trainable. 

\subsection{Map-based methods} 
Map-based methods are the mainstream of the crowd localization task. Idress~\textit{et al.}~\cite{idrees2018composition}, and Gao~\textit{et al.}~\cite{gao2019domain} utilize small Gaussian kernel density maps, and the head locations are equal to the maxima of density maps. Even though using the small kernel can generate sharp density maps, it still exists overlaps in the extremely dense region, making the head location undistinguishable. To solve this, some methods~\cite{xu2019autoscale,liang2021focal,gao2020learning,abousamra2020localization} focus on designing new maps to handle the extremely dense region, such as the distance label map~\cite{xu2019autoscale},  Focal Inverse Distance Transform Map (FIDTM)~\cite{liang2021focal} and Independent Instance Map (IIM)~\cite{gao2020learning}.  These methods can effectively avoid overlap in the dense region, but they need post-processing (``find-maxima") to extract the instance location and rely on multi-scale feature maps, which is not simple and elegant.  

\begin{figure*}[t]
\centering
\resizebox{0.97\textwidth}{!}{
    \includegraphics{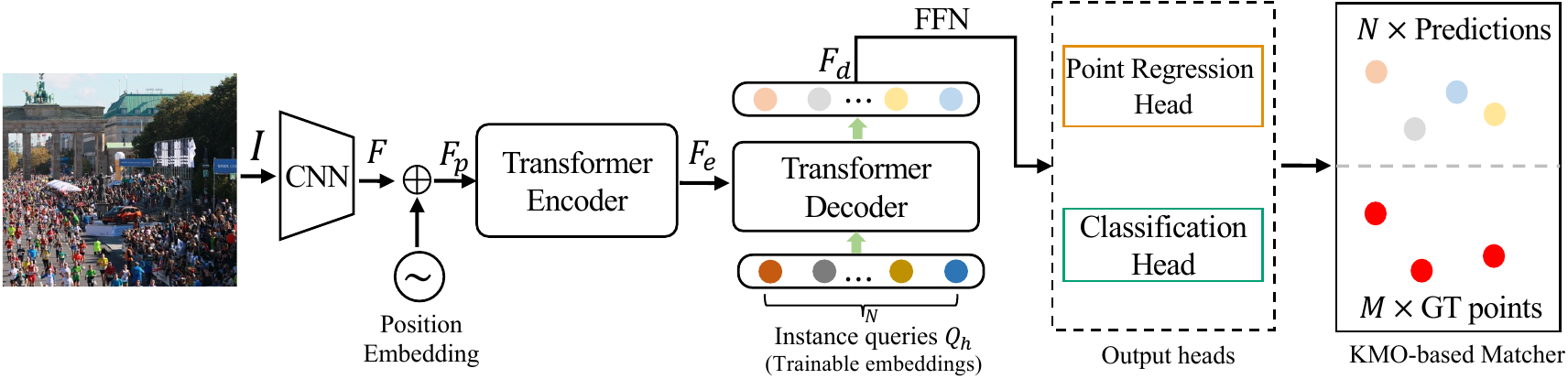}
    }
\caption{The overview of our CLTR. First, the input image $I$ is fed to the CNN-based backbone to extract the features $F$. Second, the features $F$ are added position embedding, resulting in $F_p$, fed to the transformer-encoder layers, outputting $F_e$. Third, we define $N \times$ trainable embeddings $Q_h$ as query, $F_e$ as key, and transformer decoder takes the $Q_h$ and $F_e$ as input to generate the decoded feature $F_d$. Finally, the $F_d$ can be decoupled to the point coordinate and corresponding confidence score.}
\centering
\label{fig:pipeline}
\end{figure*}

\subsection{Regression-based methods}
Just a few research works focus on regression-based. We note a recent paper~\cite{song2021rethinking}, P2PNet, also a regression-based framework for crowd localization. But this is a concurrent work that has appeared while this manuscript is under preparation. P2PNet~\cite{song2021rethinking} defines surrogate regression on a large set of proposals, and the model relies on pre-processing, such as producing $8 \times W \times H$ point proposals. In contrast, our method replaces massive fixed point proposals with a few trainable instance queries, which is more elegant and unified. 

\subsection{Visual transformer} 
Recently, visual transformers~\cite{dosovitskiy2020image,touvron2021training,liu2022end,carion2020end,meng2021conditional} have gone viral in computer vision. In particular, DETR~\cite{carion2020end} utilizes the Transformer-decoder to model object detection in an end-to-end pipeline, successfully removing the need for post-processing. Based on DETR~\cite{carion2020end}, Conditional DETR~\cite{meng2021conditional} further adopts the spatial queries and keys to a band containing the object extremity or a region, accelerating the convergence of DETR~\cite{carion2020end}. In the crowd analysis, Liang~\textit{et al.}~\cite{liang2021transcrowd,liu2021visdrone} propose TransCrowd, which reformulates the weakly-supervised counting problem from a sequence-to-count perspective. Several methods~\cite{tian2021cctrans,sun2021boosting} demonstrate the power of transformers in point-supervised crowd counting setup. Method~\cite{gao2021congested} adopts the IIM~\cite{gao2020learning} in the swin transformer~\cite{liu2021swin} to implement crowd localization.

\section{Our Method}

The overview of our method is shown in Fig.~\ref{fig:pipeline}. The proposed method is an end-to-end network, directly predicting all instances at once without additional pre-processing and post-processing. The approach consists of a CNN-based backbone, a transformer encoder, a transformer decoder, and a KMO-based matcher. Given an image by $I \in \mathbb{R}^{H \times W \times 3}$, where $H$, $W$ are the height and width of the image,
\begin{itemize}
    \item The CNN-based backbone first extracts the feature maps $F \in \mathbb{R}^{\frac{H}{32} \times \frac{W}{32} \times C}$ from the input image $I$. To verify the effectiveness of our method, the $F$ is only a single-scale feature map without feature aggregation.
    \item The feature maps $F$ are then flattened into a 1D sequence with positional embedding, and the channel dimension is reduced from $C$ to $c$, which results in $F_p \in \mathbb{R}^{\frac{HW}{32^2} \times c}$. The transformer-encoder layers take the $F_p$ as input and output encoded features $F_e$.
    \item Next, the transformer-decoder layers take the trainable head queries $Q_h$ and encoded features $F_e$ as input and interact with each other via cross attention to generate the decoded embedding $F_d$, which contains the point (person's head) and category information.
    \item Finally, the decoded embeddings $F_d$ are decoupled to the point coordinates and confidence scores by a point regression head and a classification head, respectively.
\end{itemize}

\subsection{Transformer Encoder}

We use a $1 \times 1$ convolution to reduce the channel dimension of the extracted feature maps $F$ from $\mathbb{R}^{ \frac{H}{32} \times \frac{W}{32} \times C}$ to $ \mathbb{R}^{ \frac{H}{32} \times \frac{W}{32} \times c}$ ($c$ set as 256). Due to the transformer-encoder adopt a $1D$ sequence as input, we reshape the extracted features $F$ and add position embedding, resulting in $ F_p \in \mathbb{R}^{ \frac{HW}{32^2} \times c} $. The $F_p$ are then fed into the transformer-encoder layers to generate the encoded features $F_e$. Here the encoder contains many encoder layers, and each layer consists of a self-attention ($SA$) layer and a feed-forward ($FC$) layer. The $SA$ consists of three inputs, including query ($Q$), key ($K$), and value ($V$), defined as follow:
\begin{equation}
SA (Q, K, V) = softmax(\frac{QK^T}{\sqrt{c}}) V,
\label{eq:self_attention}
\end{equation}
where $Q$, $K$ and $V$ are obtained from the same input $Z$ (\textit{e.g.,} $Q = ZW_Q$).
In particular, we use the multi-head self-attention ($MSA$) to model the complex feature relation, which is an extension with several independent $SA$ modules: $MSA = [SA_1;SA_2;\cdots;SA_m]W_{O}$, where $W_{O}$ is a re-projection matrix and $m$ is the number of attention heads set as 8.

\subsection{Transformer Decoder}

The transformer-decoder consists of many decoder layers, and each layer is composed of three sub-layers: (1) a self-attention ($SA$) layer. (b) a cross attention ($CA$) layer. (3) a feed-forward ($FC$) layer. The $SA$ and $FC$ are the same as the encoder. The $CA$ module takes two different embeddings as input instead of the same inputs in $SA$. Let us denote two embeddings as $X$ and $Z$, and the $CA$ can be written as $CA = SA (Q = XW_Q, K = ZW_K, V = ZW_V)$. Following~\cite{meng2021conditional}, we adopt the conditional cross-attention, \textit{i.e.,} the $Q$ are concatenated by the trainable embeddings $Q_h$ and content query (from decoder self-attention). The decoder output the decoded features $F_d$, which are used to predict the point coordinates (regression head) and confidence scores (classification head).

\begin{figure}[t]
	\begin{center}
		\includegraphics[width=0.96\linewidth]{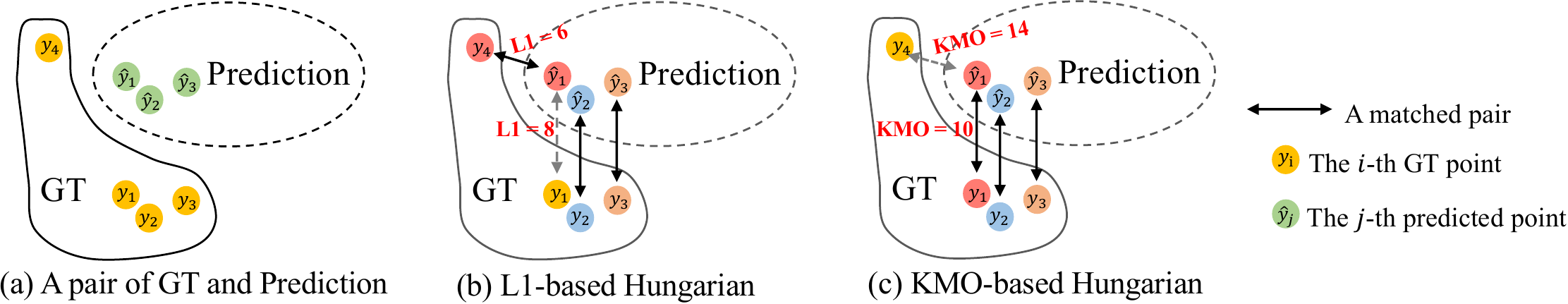}
	\end{center}
	\caption{
	(a) A pair of GT and predictions. (b) The $L1$-based Hungarian generate unsatisfactory matching results. (c) The proposed KMO-based Hungarian models the context as the matching cost, generating more reasonable matching results.
	}
	\label{fig:kmo}
\end{figure}

\subsection{KMO-based Matcher}
To train the model, we need to match the predictions and GT by one-to-one, and the unmatched predicted points are considered to the ``background" class.

Let us denote the predicted points set as $\hat{y} = \{\hat{y}_j\}_{j=1}^{N}$ and GT points set as $y = \{y_i\}_{i=1}^{M}$. $N$ and $M$ refer to the number of predicted heads and GT, respectively. $N$ is larger than $M$ to ensure each GT matches a prediction, and the rest of the predictions match failed can be classified as negative. Next, we need to find a bipartite matching between these two sets with the lowest cost. A straightforward way is to take the $L1$ distance and confidence as matching cost:
\begin{equation}
\begin{aligned}
L_{m} (y_i, \hat{y}_j) = ||y^p_i - \hat{y}_j^p||_1 - \hat{C_j}, i \in M, j \in N, 
\label{eq:l2_hungarian}
\end{aligned}
\end{equation}
where $||*||_1$ means the $L1$ distance and $\hat{C_j}$ is the confidence of the $j$-th predicted point. $y^p_i$ is a vector that defines the $i$-th GT point coordinates. Accordingly, $\hat{y}_j^p $ is formed as the point coordinates of $j$-th predicted head. Based on the $L_{m}$, we can utilize the Hungarian~\cite{kuhn1955hungarian} to implement one-to-one matching. However, we find that merely taking the $L1$ with confidence maybe generate unsatisfactory matching results (seen Fig.~\ref{fig:intro}(d)). Another toy example is shown in Fig.~\ref{fig:kmo}, given a pair of GT and prediction set (Fig.~\ref{fig:kmo}(a)), from the whole perspective, the $\hat{y}_1$ should match the $y_1$ ideally (just like $\hat{y}_2$ match $y_2$). 
Using Eq.~\ref{eq:l2_hungarian} for matching cost\footnote{Here, we ignore the $\hat{C_j}$ for simply illustrating since heads usually report similar confidence score.}, it will match the $\hat{y}_1$ and $y_4$ since the $L1$-based Hungarian lack of context information.
Thus, we introduce the KMO-based Hungarian, which effectively utilizes the context as auxiliary matching cost, formulated as $L_m^k$:
\begin{equation}
\begin{gathered}
L_m^k (y_i, \hat{y}_j) = ||y^p_i - \hat{y}_j^p||_1 + ||y^k_i - \hat{y}_j^k||_1 - \hat{C_j},
\\
y^k_i = \frac{1}{k}\sum_{k=1}^{k} d_i^{k},\qquad \hat{y}_j^k =  \frac{1}{k}\sum_{k=1}^{k} \hat{d}_j^{k},
\label{eq:kmo_hungarian_1}
\end{gathered}
\end{equation}
where $d^k_i$ means the distance between $i$-th GT point and its $k$-th neighbour. $y^k_i$ refer to the average neighbour-distance of the $i$-th GT point. $\hat{d}^k_j$ and $\hat{y}^k_j$ have similar definitions as $d^k_i$ and $y^k_i$, respectively. Taking inspirations from~\cite{liu2019point,sam2020locate}, in our experiments, $\hat{y}^k_j$ is predicted by the network. 
The proposed $L_m^k$, revisiting the label assignment from a context view, turns to find the whole optimum. As shown in Fig.~\ref{fig:kmo}(c), the proposed KMO-based Hungarian makes sure the $\hat{y}_1$ successfully matches the $y_1$. 
Regarding the predictions as a point set containing the geometric relationships. Assignment on Fig.~\ref{fig:kmo}(b) is not wrong, but it is a local view without context, and it will break the internal geometric relationships of the point set. Assignment on Fig.~\ref{fig:kmo}(c) considers the context information from the nearby heads, pursuing the whole optimum and maintaining the geometric relationships of the point set, making the model easier to be optimized, which is more reasonable.
For the case in Fig.~\ref{fig:intro}(d), when multiple \textit{gts} tend to match the same predicted point, the KMO-based Hungarian will resolve their conflicts by using the context information. Note that the matcher is just used in the training phase.

\subsection{Loss function}
After obtaining the one-to-one matching results, we calculate the loss for back-propagate. We make point predictions directly. The loss consists of point regression and classification. For the point regression, we employ the commonly-used $L_1$ loss, defined as:
\begin{equation}
\begin{aligned}
L_{loc} = ||y_i^p - \hat{y}_{\sigma (i)}^p||_1, 
\label{eq:kmo_hungarian_2}
\end{aligned}
\end{equation}
where $\hat{y}_{\sigma (i)}^p$ is the matched subset from $y_i^p$ by using the proposed KMO-based Hungarian. It is noteworthy that we normalize all ground truth point range to $[0, 1]$ for scale invariance. We utilize the focal loss as the classification loss $L_{cls}$, and the final loss $L$ is defined as:
\begin{equation}
\begin{aligned}
L = L_{cls} + \lambda L_{loc},
\label{eq:kmo_hungarian_3}
\end{aligned}
\end{equation}
where $\lambda$ is a balance weight, set as 2.5. These two losses are normalized by the number of instances inside the batch.

\section{Experiments}

\subsection{Implementation details}  
We use the ResNet50~\cite{he2016deep} as the backbone. The number of transformer encoder layers and decoder layers are both set to 6. The $N$ is set to 500 (number of instance queries $Q_h$). We augment the training data using random cropping, random scaling, and horizontal ﬂipping with a 0.5 probability. The crop size is set as $128 \times 128$ for ShanghaiTech Part A, $256 \times 256$ for the rest datasets. We use Adam with the learning rate of 1e-4 to optimize the model. For the large-scale datasets (\textit{i.e.}, UCF-QNRF, JHU-Crowd++, and NWPU-Crowd), we ensure the longer size is less than 2048, keeping the original aspect ratio. The batch size is set to 16. $k$ is set as 4 for all datasets. During the testing phase, each image is split into non-overlapped patches (size same as training phase). Zero padding is adopted if a cropped patch is smaller than the predefined size. And a simple confidence threshold (set to $0.35$) is used to filter the ``background" class. 

\begin{table*}[t]
\scriptsize
\centering
\setlength{\tabcolsep}{3mm}
\caption{Localization performance on NWPU-Crowd dataset. * means the methods rely on box-level instead of point-level annotations.}
\begin{tabular}{ccccccc}
 \toprule
{\multirow{2}{*}{Method}} &\multicolumn{3}{c}{Validation set} &\multicolumn{3}{c}{Test set} \\ 
\cmidrule{2-7}
&P(\%) &R(\%) &F(\%) &P(\%) &R(\%) &F(\%) \\
\midrule
Faster RCNN*~\cite{ren2015faster} & \textbf{96.4}\% & 3.8\% & 7.3\% & \textbf{95.8}\% & 3.5\% & 6.7\% \\
TinyFaces*~\cite{hu2017finding} & 54.3\% & \textbf{66.6}\% & \textbf{59.8}\% & 52.9\% & 61.1\% & 56.7\% \\
TopoCount*~\cite{abousamra2020localization} & - & - & - & 69.5\% & \textbf{68.7}\% & \textbf{69.1}\% \\
\midrule
GPR~\cite{gao2019domain} & 61.0\% & 52.2\% & 56.3\% & 55.8\% & 49.6\% & 52.5\% \\
RAZ\_Loc~\cite{liu2019recurrent} & 69.2\% & 56.9\% & 62.5\% & 66.6\% & 54.3\% & 59.8\% \\
AutoScale\_loc~\cite{xu2019autoscale} & 70.1\% & 63.8\% & 66.8\% & 67.3\% & 57.4\% & 62.0\% \\
Crowd-SDNet~\cite{wang2021self} &-&-&-&65.1\%&62.4\%&63.7\%\\
GL~\cite{wan2021generalized} & - & - & - & \textbf{80.0}\% & 56.2\% & 66.0\% \\
CLTR (\textbf{ours}) & \textbf{73.9}\% & \textbf{71.3}\% & \textbf{72.6}\% & 69.4\% & \textbf{67.6}\% & \textbf{68.5}\% \\
\bottomrule
\end{tabular}
\label{tab:NWPU_loc}
\end{table*}

\begin{table*}[t]
\scriptsize
\centering
\setlength{\tabcolsep}{4mm}
\caption{Localization performance on the UCF-QNRF dataset. We report the Average Precision, Recall, and F1-measure at different thresholds $\sigma$: $(1, 2, 3, \dots, 100)$ pixels.
}
\begin{tabular}{cccc}
\toprule
Method &Av.Precision&Av.Recall&F1-measure\\
\midrule
CL~\cite{idrees2018composition}&75.80\%&59.75\%&66.82\%\\
LCFCN~\cite{laradji2018blobs}&\textcolor{black}{77.89\%}&\textcolor{black}{52.40\%}&\textcolor{black}{62.65\%}\\
Method in~\cite{ribera2019}&\textcolor{black}{75.46\%}&\textcolor{black}{49.87\%} & \textcolor{black}{60.05\%}\\
LSC-CNN\cite{sam2020locate} &75.84\% &74.69\% & 75.26\%\\
AutoScale\_loc~\cite{xu2019autoscale} & 81.31\%&75.75\%&78.43\%\\
GL~\cite{wan2021generalized} & 78.20\%&74.80\%& 76.40\%\\
TopoCount~\cite{abousamra2020localization} &81.77\% &78.96\% & \textcolor{black}{80.34\%}\\
CLTR \textbf{(ours)}&\textbf{82.22}\%&\textbf{79.75}\% &\textbf{80.97}\%\\
\bottomrule
\end{tabular}
\label{tab:qnrf_loc}
\end{table*}

\begin{table*}[t]
\scriptsize
\centering
\setlength{\tabcolsep}{3mm}
\caption{Comparison of the localization performance on the Part A dataset.}
\begin{tabular}{ ccccccc }
 \toprule
 {\multirow{2}{*}{Method}}  &\multicolumn{3}{c}{$\sigma$ = 4}&\multicolumn{3}{c}{$\sigma$ = 8}  \\
\cmidrule{2-7}
& P (\%)& R (\%) &F (\%) &P (\%) &R (\%)&F (\%)  \\
\midrule
LCFCN\cite{laradji2018blobs}  &43.3\% &26.0\% &32.5\%& \textbf{75.1}\% &45.1\% &56.3\% \\
Method in~\cite{ribera2019}  &34.9\% &20.7\% &25.9\%& 67.7\% &44.8\%& 53.9\% \\
LSC-CNN~\cite{sam2020locate}  &33.4\% &31.9\% & 32.6\%&63.9\% &61.0\% &62.4\%  \\
TopoCount~\cite{abousamra2020localization} &41.7\% &40.6\% &41.1\% &74.6\% &72.7\% &73.6\% \\
CLTR \textbf{(ours)} &\textbf{43.6}\%&\textbf{42.7}\%&\textbf{43.2}\%&74.9\%&\textbf{73.5}\% &\textbf{74.2}\%
\\
\bottomrule
\end{tabular}
\label{tab:A_loc}
\end{table*}

\subsection{Dataset} 
We evaluate our method on five challenging public datasets, each being elaborated below. 

\textbf{NWPU-Crowd}~\cite{gao2020nwpu} is a large-scale dataset collected from various scenes, consisting of 5,109 images. The images are randomly split into training, validation, and test sets, which contain 3109, 500, and 1500 images, respectively. This dataset provides point-level and box-level annotations. 

\textbf{JHU-Crowd++}~\cite{sindagi2020jhu-crowd++} is a challenging dataset containing 4372 crowd images. This dataset consists of 2272 training images, 500 validation images, and 1600 test images. And the total number of people in each image ranges from 0 to 25791.

\textbf{UCF-QNRF}~\cite{idrees2018composition}, a dense dataset, contains 1535 images (1201 for training and 334 for testing) and about one million annotations. The average number of pedestrians per image is 815, and the maximum number reaches 12865.

\textbf{ShanghaiTech}~\cite{zhang2016single} is divided into Part A and Part B. Part A consists of 300 training images and 182 test images. Part B consists of 400 training images and 316 test images. 

\subsection{Evaluation Metrics}
This paper mainly focuses on crowd localization, and counting is an incidental task,~\textit{i.e.,} the total count is equal to the number of predicted points.

\textbf{Localization Metrics.}
In this work, we utilize the Precision, Recall, and F1-measure as the localization metrics, following ~\cite{gao2020nwpu,idrees2018composition}. If the distance between a predicted point and GT point is less than the predefined distance threshold $\sigma$, this predicted point will be treated as True Positive (TP). For the NWPU-Crowd dataset~\cite{gao2020nwpu}, containing the box-level annotations, we set $\sigma$ to ${\sqrt {{w^2} + {h^2}}}/2 $, where $w$ and $h$ are the width and height of each head, respectively. For the ShanghaiTech dataset, we utilize two fixed thresholds, including $\sigma = 4$ and $\sigma = 8$. For the UCF-QNRF, we use various threshold ranges from [1, 100], following CL~\cite{idrees2018composition}.

\textbf{Counting Metrics.}
The Mean Absolute Error (MAE) and Mean Square Error (MSE) are used as counting metrics, defined as: $MAE=\frac{1}{N_c}\sum_{i=1}^{N_c}\left |P_{i}-G_{i} \right|$, $MSE=\sqrt{\frac{1}{N_c}\sum_{i=1}^{N_c}\left |P_{i}-G_{i} \right|^{2}}$, where $N_c$ is the total number of images, $P_{i}$ and $G_{i}$ are the predicted and GT count of the $i$-th image, respectively. 

\section{Results and Analysis}

\subsection{Crowd Localization}
We first evaluate the localization performance with some state-of-the-art localization methods~\cite{wan2021generalized,abousamra2020localization,xu2019autoscale}, as shown in Table~\ref{tab:NWPU_loc}, Table~\ref{tab:qnrf_loc}, and Table~\ref{tab:A_loc}. For the NWPU-Crowd (see Table~\ref{tab:NWPU_loc}), a large-scale dataset, our CLTR outperforms GL~\cite{wan2021generalized} and AutoScale~\cite{xu2019autoscale} at least 5.8\% (\textit{resp.} 2.5\%) for F1-measure on the validation set (\textit{resp.} test set). It is noteworthy that this dataset provides precise box-level annotations, and the TopoCount~\cite{abousamra2020localization} relies on the labeled box in the training phase instead of using the point-level annotations. Even though our method is just based on point-annotation, a more weak label mechanism, it can still achieve competitive performance against the TopoCount~\cite{abousamra2020localization} on the NWPU-Crowd (test set). For the dense dataset, UCF-QNRF (see Table~\ref{tab:qnrf_loc}), our method achieves the best Average Precision, Average Recall and F1-measure. For the ShanghaiTech Part A (see Table~\ref{tab:A_loc}), a sparse dataset, our CLTR outperforms the state-of-the-art method TopoCount~\cite{abousamra2020localization} by 2.1\% F1-measure for the strict setting ($\sigma =4$), and still get ahead for the less strict settings ($\sigma = 8$). These results demonstrate that the proposed method can cope with various scenes, including large-scale, dense and sparse scenes. Note that all of other localization methods~\cite{xu2019autoscale,wan2021generalized,abousamra2020localization} adopt multi-scale or higher-resolution feature that potentially benefit our approach, which is currently not our focus and left as our future work.

\begin{table}[t]
\scriptsize
\centering
\setlength{\tabcolsep}{4.1mm}
\caption{Counting results of various methods on the NWPU validation and test sets.}
\resizebox{0.9\textwidth}{!}{
\begin{tabular}{cccccc}
\toprule
{\multirow{3}{*}{Method}} &{\multirow{3}{*}{\tabincell{c}{Output\\Position\\Information}}} &\multicolumn{2}{c}{\multirow{2}{*}{Validation set}} &\multicolumn{2}{c}{\multirow{2}{*}{Test set}} \\ 
&&\multicolumn{2}{c}{}&\multicolumn{2}{c}{}\\
\cmidrule{3-6}
&&MAE&MSE&MAE&MSE \\
\midrule
MCNN~\cite{zhang2016single} &\xmark &218.5 &700.6 &232.5 &714.6 \\
CSRNet~\cite{li2018csrnet} &\xmark &104.8 &433.4 &121.3 &387.8 \\
CAN~\cite{liu2019context} &\xmark &93.5 &489.9 &106.3 &\textbf{386.5} \\
SCAR~\cite{gao2019scar}&\xmark  &81.5 &397.9 &110.0 &495.3 \\
BL~\cite{ma2019bayesian} &\xmark &93.6 &470.3 &105.4 &454.2 \\
SFCN~\cite{wang2019learning} &\xmark &95.4 &608.3 &105.4 &424.1 \\
DM-Count~\cite{wang2020distribution} &\xmark &\textbf{70.5} &\textbf{357.6} &\textbf{88.4} &388.6 \\
\midrule
RAZ\_loc~\cite{liu2019recurrent} &\rmark &128.7&665.4 & 151.4 &634.6 \\
AutoScale\_loc~\cite{xu2019autoscale} &\rmark &97.3&571.2&123.9&515.5 \\
TopoCount~\cite{abousamra2020localization} &\rmark &-&-&107.8&438.5 \\
GL~\cite{wan2021generalized} &\rmark &-&-&79.3&346.1\\
CLTR (\textbf{ours}) &\rmark & \textbf{61.9}& \textbf{246.3}& \textbf{74.3}&\textbf{333.8} \\
\bottomrule
\end{tabular}}
\label{tab:NWPU_count}
\end{table}

\begin{table*}[!t]
\scriptsize
\centering
\setlength{\tabcolsep}{1.5mm}
\caption{Comparison of the counting performance on the UCF-QNRF, ShanghaiTech Part A, and Part B datasets.}
\resizebox{0.8\textwidth}{!}{
\begin{tabular}{lccccccc}
 \toprule
 {\multirow{3}{*}{Method}} &{\multirow{3}{*}{\tabincell{c}{Output\\Position\\Information}}}&\multicolumn{2}{c}{\multirow{2}*{QNRF}}   &\multicolumn{2}{c}{\multirow{2}*{Part A}} &\multicolumn{2}{c}{\multirow{2}*{Part B}} \\
&&&&&&&\\
\cmidrule{3-8}
&& MAE & MSE&MAE&MSE &MAE&MSE \\
\midrule
     CSRNet~\cite{li2018csrnet}                 &\xmark   &-&-            &68.2&115.0   & 10.6&16.0 \\
     L2SM~\cite{xu2019learn}                     &\xmark  &104.7 &173.6  &64.2&98.4   &7.2 &11.1\\ 
     DSSI-Net \cite{liu2019crowd}&\xmark &99.1&159.2&60.6&96.0&6.9&10.3\\
     MBTTBF \cite{sindagi2019multi} &\xmark &97.5&165.2&60.2&94.1&8.0&15.5\\
     BL~\cite{ma2019bayesian}  &\xmark &88.7&154.8                       &62.8&101.8   & 7.7 & 12.7\\
     AMSNet~\cite{hu2020count} &\xmark &101.8&163.2                  &56.7&\textbf{93.4}    &\textbf{6.7}&\textbf{10.2}\\
     LibraNet~\cite{liu2020weighing} &\xmark  &88.1 & \textbf{143.7}&\textbf{55.9}&97.1      &7.3&11.3\\
    KDMG~\cite{wan2020kernel}   &\xmark &99.5&173.0&63.8&99.2&7.8&12.7\\
     NoisyCC~\cite{wan2020modeling}   &\xmark  &85.8&150.6&61.9&99.6&7.4&11.3\\
     DM-Count ~\cite{wang2020distribution}  &\xmark &\textbf{85.6}&148.3&59.7&95.7&7.4&11.8 \\
     \midrule
     CL~\cite{idrees2018composition}    &\rmark  &132.0&191.0&-&-&-&-\\
     RAZ\_loc+~\cite{liu2019recurrent} &\rmark   &118.0&198.0  &71.6& 120.1 & 9.9 & 15.6  \\
     PSDDN~\cite{liu2019point}                 &\rmark    &-&- &65.9&112.3&9.1&14.2\\
     LSC-CNN~\cite{sam2020locate} &\rmark   &120.5&218.2 &66.4&117.0  &8.1 &12.7\\
     TopoCount~\cite{abousamra2020localization}  &\rmark   &89.0&159.0&61.2&104.6&7.8&13.7 \\
     AutoScale\_loc~\cite{xu2019autoscale} &\rmark &104.4&174.2&65.8&112.1&8.6&13.9 \\
     GL~\cite{wan2021generalized}  &\rmark  &\textbf{84.3}&147.5&61.3&95.4&7.3&11.7 \\
     CLTR \textbf{(ours)}&\rmark &85.8&\textbf{141.3}& \textbf{56.9}&\textbf{95.2} &\textbf{6.5}&\textbf{10.6} \\
     \bottomrule
\end{tabular}}
\label{tab:qnrf_A_B_counting}
\end{table*}

\begin{table*}[t]
\scriptsize
\centering
\setlength{\tabcolsep}{0.5mm}
\caption{Categorical counting results on JHU-Crowd++ dataset. Low, Medium, and High respectively indicate three categories based on different ranges: [0, 50], (50, 500], and (500, $+\infty$). Weather means the degraded images (\textit{e.g.,} haze, snow, rain).}
\begin{tabular}{ lccccccccccc }
\toprule
 \multirow{3}{*}{Methods}&{\multirow{3}{*}{\tabincell{c}{Output\\Position\\Information}}}&\multicolumn{2}{c}{\multirow{2}*{Low}} &\multicolumn{2}{c}{\multirow{2}*{Medium}} &\multicolumn{2}{c}{\multirow{2}*{High}} &\multicolumn{2}{c}{\multirow{2}*{Weather}} &\multicolumn{2}{c}{\multirow{2}*{Overall}} \\
 &&\multicolumn{2}{c}{}&\multicolumn{2}{c}{}&\multicolumn{2}{c}{}&\multicolumn{2}{c}{}&\multicolumn{2}{c}{}\\
 \cmidrule{3-12}
 && MAE & MSE & MAE & MSE & MAE & MSE & MAE & MSE & MAE & MSE \\
 \midrule
 MCNN~\cite{zhang2016single} &\xmark&97.1 & 192.3& 121.4& 191.3& 618.6& 1,166.7 &330.6 & 852.1& 188.9& 483.4  \\
 CSRNET~\cite{li2018csrnet} &\xmark& 27.1& 64.9& 43.9& 71.2& 356.2& 784.4 &141.4 & 640.1& 85.9& 309.2\\
 JHU++~\cite{sindagi2020jhu-crowd++} &\xmark& 14.0 & 42.8 & 35.0 & \textbf{53.7} & \textbf{314.7} & \textbf{712.3} & \textbf{120.0} & \textbf{580.8} & \textbf{71.0} & \textbf{278.6}\\
 LSC-CNN~\cite{sam2020locate}  &\xmark&10.6 & \textbf{31.8} &34.9 & 55.6& 601.9& 1,172.2 & 178.0 & 744.3& 112.7 & 454.4 \\
 BL~\cite{ma2019bayesian} &\xmark&\textbf{10.1}&32.7& \textbf{34.2}&54.5& 352.0& 768.7 & 140.1 & 675.7&75.0& 299.9\\
\midrule
 AutoScale\_loc~\cite{xu2019autoscale} &\rmark &13.2&30.2&32.3&52.8&425.6&916.5&-&-& 85.6& 356.1 \\
  TopoCount~\cite{abousamra2020localization}&\rmark &\textbf{8.2}&\textbf{20.5}&\textbf{28.9}&\textbf{50.0}&282.0&685.8&120.4&635.1&60.9&267.4\\
 GL~\cite{wan2021generalized} &\rmark&-&-&-&-&-&-&-&-&59.9&259.5\\
 CLTR (\textbf{ours}) &\rmark &8.3&21.8&30.7&53.8&\textbf{265.2}&\textbf{614.0}&\textbf{109.5}&\textbf{568.5}&\textbf{59.5}&\textbf{240.6}\\
 \bottomrule
 \end{tabular}
\label{tab:jhu_count}
\end{table*}

\subsection{Crowd Counting}
In this section, we compare the counting performance with various methods (including density map regression-based and localization-based), as shown in Table~\ref{tab:NWPU_count}, Table~\ref{tab:qnrf_A_B_counting} and Table~\ref{tab:jhu_count}. Although our method only uses a single-scale and low-resolution ($\frac{1}{32}$ of input image) feature map, it can achieve state-of-the-art or highly competitive performance in all experiments. Specifically, our method achieves the first MAE and MSE on the NWPU-Crowd test set (see Table~\ref{tab:NWPU_count}). Compared with the localization-based counting methods (the bottom part of Table~\ref{tab:qnrf_A_B_counting}), which can provide the position information, our method achieves the best counting performance in MAE and MSE on ShanghaiTech Part A and Part B datasets. On the UCF-QNRF dataset, our method achieves the best MSE and reports comparable MAE. On the JHU-Crowd++ dataset ( Table~\ref{tab:jhu_count}), our method outperforms the state-of-the-art method GL~\cite{wan2021generalized} by a signiﬁcant margin of 18.9 MSE. Furthermore, CLTR has superior performance on the extremely dense (the ``High" part) and degraded set (the ``Weather" part).  

\begin{figure*}[ht]
\centering
\resizebox{0.97\textwidth}{!}{
    \includegraphics{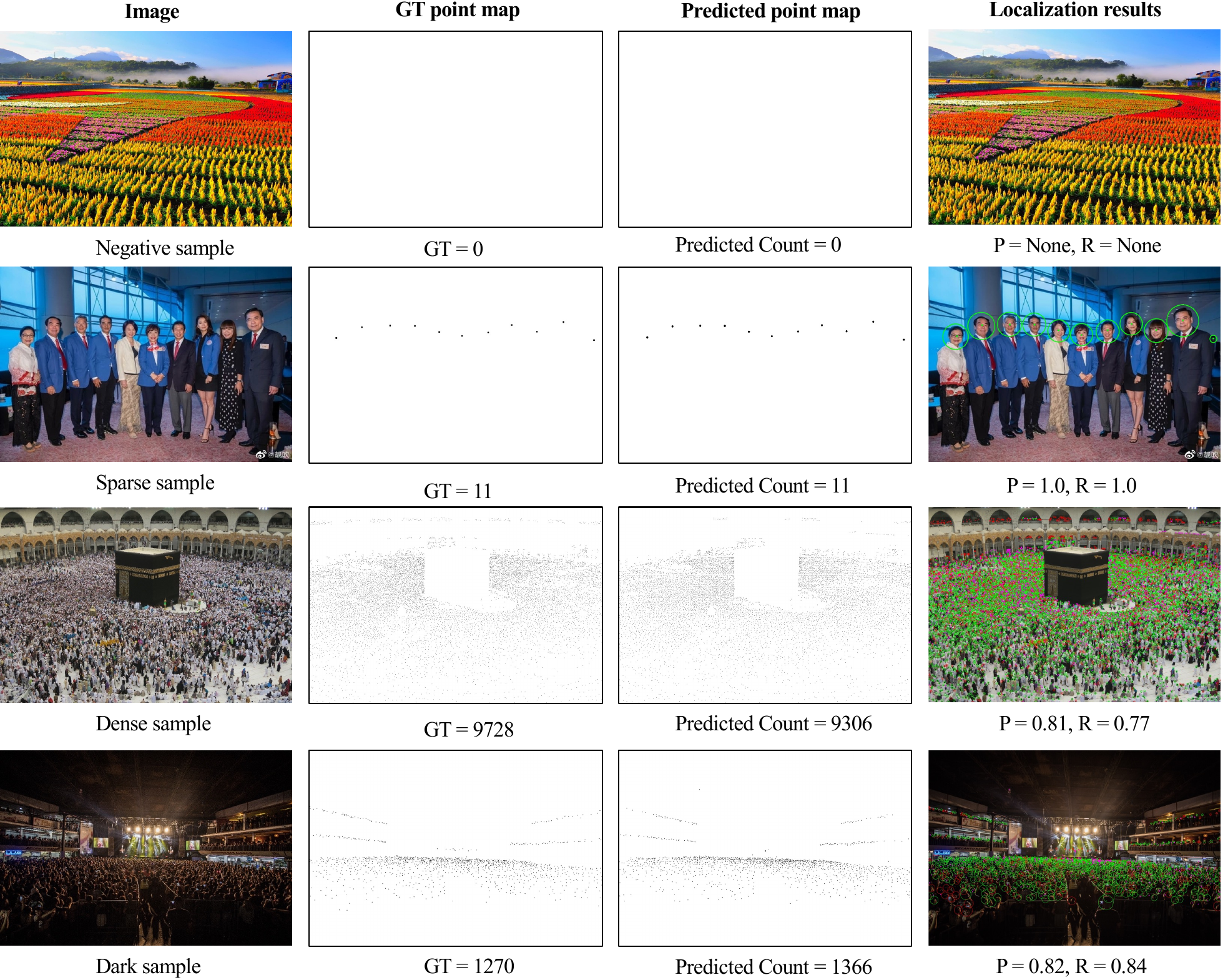}
    }
\caption{Some examples from the NWPU-Crowd dataset (validation set). From left to right, there are images, GT points, predicted points, and localization results. Row 1, row 2, row 3, and row 4  refer to the negative, sparse, dense and dark samples, respectively. In the last column, $P$ and $R$ are the Precision and Recall, respectively. The \textcolor{green}{green}, \textcolor{red}{red} and \textcolor{magenta}{magenta} points denote true positive (TP), false negative (FN) and false positive (FP), respectively. It can be seen that the proposed method can effectively handle various scenes.}
\centering
\label{fig:visiual}
\end{figure*}

\begin{table}[t]
\scriptsize
\centering
\setlength{\tabcolsep}{3mm}
\caption{Effect of transformer size (the number of layers and the number of trainable queries $Q_h$) on UCF-QNRF dataset.}
\begin{tabular}{ccccccc}
\toprule
 \multirow{2}{*}{ Layers} &\multirow{2}{*}{\tabincell{c}{$N$\\(Queries number)}} &\multicolumn{3}{c}{Localization} &\multicolumn{2}{c}{Counting}\\
\cmidrule{3-7}
 & & P (\%) & R (\%)& F (\%)& MAE & MSE\\
\midrule
 3  & 500 & 80.60\% & 79.44\% &80.02\%&88.4&149.9\\
 6  & 500 & \textbf{82.22}\% & \textbf{79.75}\% &\textbf{80.97}\%&\textbf{85.8}&\textbf{141.3}\\
 12 & 500 & 80.82\% & 79.41\%& 80.11\%& 87.7&150.3\\
 \midrule
 6  & 300 & 80.61\% & 79.18\% &79.89\%&89.9&153.6\\
 6 &  700 & 81.32\% & 79.38\% &80.34\%&86.8&146.4\\
 \bottomrule
\end{tabular}
\label{tab:influence_transformer}
\end{table}

\subsection{Visualizations}
We further give some qualitative visualizations to analyze the effectiveness of our method, as shown in Fig.~\ref{fig:visiual}. The samples are selected from some typical scenes on the NWPU-Crowd dataset (validation set), including negative, sparse, extremely dense and dark scenes. In the first row, CLTR shows a strong robust on the negative sample (“dense fake humans”). CLTR performs well in different congested scenes, such as the sparse scene (row 2) and extremely dense scene (row 3). Additionally, we find that CLTR can also make promising localization results in dark scenes (row 4). These impressive visualizations demonstrate the effectiveness of our method in crowd localization and counting. 

\subsection{Ablation studies}
The ablation studies are carried out on the UCF-QNRF dataset, a large and dense dataset, which can effectively avoid overfitting. 

\subsubsection{Effect of transformer.}

We first study the influence by changing the size of the transformer, including the number of encoder/decoder layers and trainable instance queries. As listed in Table~\ref{tab:influence_transformer}, we find that when the layer and queries number are set to 6 and 500, the CLTR achieves the best performance. When the number of queries changes to 700 (resp. 300), the performance of MAE drops from 85.8 to 86.8 (resp. 89.9). We hypothesize that, by using a small number of queries, CLTR may lose potential heads, while using a large number of queries, CLTR may generate massive negative samples. We empirically find that all the pre-defined non-overlap patches contain less than 500 persons. The following ablation studies are organized using 6 transformer layers and 500 queries.

\begin{table*}[t]
\scriptsize
\centering
\setlength{\tabcolsep}{1.6mm}
\caption{The effectiveness of the proposed KMO-based Hungarian on UCF-QNRF dataset. $L_m$ only adopts the $L1$ distance with confidence as matching cost, and $L_m^k$ contains the proposed KMO.}
\begin{tabular}{ cccccc }
\toprule
 \multirow{2}{*}{Matching cost}&\multicolumn{3}{c}{Localization}&\multicolumn{2}{c}{Counting}\\
 \cmidrule{2-6}
 & Av.Precision(\%) & Av.Recall(\%)& F1-measure(\%) &MAE & MSE\\
\midrule
 $L_m$  &80.89\% & 79.17\% & 80.02\% & 91.3 & 157.4 \\
 $L_m^k$ (\textbf{ours})  & \textbf{82.22}\% & \textbf{79.75}\% & \textbf{80.97}\% & \textbf{85.8} & \textbf{141.3}  \\
 \bottomrule
\end{tabular}
\label{tab:influence_kmo}
\end{table*}

\begin{table*}[t]
\scriptsize
\centering
\setlength{\tabcolsep}{4mm}
\caption{The influence of using different numbers of nearest-neighbour on the UCF-QNRF dataset.}
\begin{tabular}{ cccccc }
\toprule
 \multirow{2}{*}{$k$}&\multicolumn{3}{c}{Localization}&\multicolumn{2}{c}{Counting}\\
 \cmidrule{2-6}
 &Av.Precision(\%) & Av.Recall(\%)& F1-measure(\%)  &MAE & MSE\\
\midrule
 3  & 81.46\%  & 79.19\% & 80.31\% & 87.1 & 146.8 \\
 4  & \textbf{82.22}\%  & \textbf{79.75}\% & \textbf{80.97}\% & \textbf{85.8} & \textbf{141.3} \\
 5  & 81.52\%  & 79.34\% & 80.42\% & 86.9 & 148.1 \\
 \bottomrule
\end{tabular}
\label{tab:influence_k}
\end{table*}

\subsubsection{Effect of matching cost.}
We next study the impact of the proposed KMO, as shown in Table~\ref{tab:influence_kmo}. When removing the KMO, we observe a significant performance drop for the counting (MAE from 85.8 to 91.3) and localization as well. We hypothesize that the $L1$ with classification can not provide a strong matching indicator, while the proposed KMO gives a direct signal to achieve great one-to-one matching based on whole-optimal. 

\begin{figure}[t]
	\begin{center}
		\includegraphics[width=0.96\linewidth]{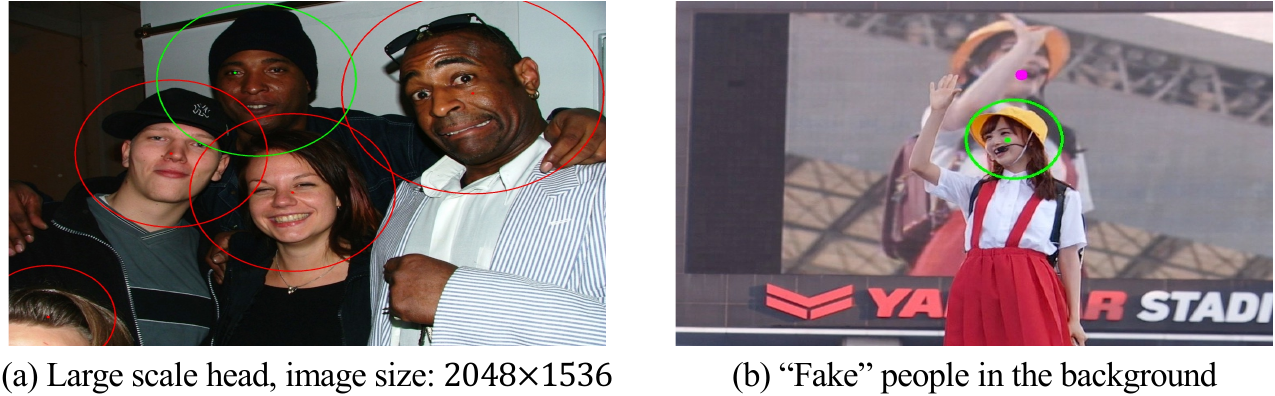}
	\end{center}
	\caption{Failure cases analysis. (a) Large-scale head, significantly larger than the crop-size. (b) Confusing regions that do not need to localize. The \textcolor{green}{green}, \textcolor{red}{red}, and \textcolor{magenta}{magenta} denote true positive (TP), false negative (FN), and false positive (FP), respectively.}
	\label{fig:failure_case}
\end{figure}

\begin{table}[t]
\scriptsize
\centering
\setlength{\tabcolsep}{4mm}
\caption{The comparisons of complexity. The experiments are conducted on a 3090 GPU, and the size of the input image is 1024 $\times$ 768. 
}
\begin{tabular}{cccc}
\toprule
Method  &Parameters (M)& MACs (G)\\
\midrule
LSC-CNN~\cite{sam2020locate} & 35.0 & 1244.3\\
AutoScale~\cite{xu2019autoscale} & 24.9 &1074.6\\
TopoCount~\cite{abousamra2020localization}& 25.8&797.2\\
GL~\cite{wan2021generalized} &\textbf{21.5} &324.6\\
CLTR (\textbf{ours}) & 43.4  &\textbf{157.2}\\
\bottomrule
\end{tabular}
\label{tab:run_time}
\end{table}

\subsubsection{Effect of $K$.}
We then study the effect of using different $k$ (the number of nearest-neighbor), listed in Table~\ref{tab:influence_k}. The proposed CLTR with different $k$ consistently achieves improvement compared with the baseline, demonstrating the proposed KMO-based Hungarian's effectiveness. When the $k$ is set to 4, we find that the result achieves the best on the UCF-QNRF dataset. We then set the same $k$ in all datasets without further fine-tuning, which works well. We also try to use a fixed radius around each point and take as many NN as they fall within that circle. However, the training time is unacceptable because calculating dynamic KNN in each circle is time-consuming.

\subsubsection{The computational statistics.}
Finally, we report the Multiply-Accumulate Operations (MACs) and parameters, as listed in Table~\ref{tab:run_time}. 
Although the proposed method has the largest parameters (mainly from the transformer part), it still reports the smallest MACs. Speeding up our model is a future work that is worthy of being studied. 

\subsection{Limitations}

Our method has some limitations. For instance, due to the CLTR crop a fixed-size (\textit{i.e.,} $256 \times 256$) sub-image for training and testing, it may fail on extremely large heads, which are significantly larger than the crop size, as shown in Fig.~\ref{fig:failure_case}(a). This problem can be solved by resizing the image into a small resolution. Another case of unsatisfied localization is shown in Fig.~\ref{fig:failure_case}(b), where there are some confused background regions (containing ``fake" people that do not need localization). 
This failure case can be solved using more modalities, such as thermal images. 


\section{Conclusion}
In this work, we propose an end-to-end crowd localization framework named CLTR, solving the task in the regression-based paradigm. The proposed method follows a one-to-one matching mechanism during the training phase. To achieve a good matching result, we propose the KMO-based Hungarian matcher, using the context information as an auxiliary matching cost. Our approach is simple yet effective. Experiments on five challenge datasets demonstrate the effectiveness of our methods. We hope our method can provide a new perspective for the crowd localization task.

\subsection*{Acknowledgment}
This work was supported by National Key R\&D Program of China (Grant No. 2018YFB1004602).

\clearpage
%
%
\bibliographystyle{splncs04}
\bibliography{egbib}
\end{document}